\documentclass[letterpaper, 10 pt, conference]{ieeeconf}

\usepackage{cite}

\usepackage[pdftex]{graphicx}
\usepackage{subcaption}
\usepackage{multirow}

\usepackage{pgfplots}
\pgfplotsset{width=7cm,compat=newest}

\usepackage{multirow}

\usepackage{amsmath}
\usepackage{units}
\usepackage[nolist]{acronym}
\usepackage{array}
\usepackage{balance}

\newcommand{\disable}[1]{}

\definecolor{dai_ligth_grey}{RGB}{230,230,230}
\definecolor{dai_ligth_grey20K}{RGB}{200,200,200}
\definecolor{dai_ligth_grey40K}{RGB}{158,158,158}
\definecolor{dai_ligth_grey60K}{RGB}{112,112,112}
\definecolor{dai_ligth_grey80K}{RGB}{68,68,68}

\definecolor{dai_petrol}{RGB}{0,103,127}
\definecolor{dai_petrol20K}{RGB}{0,86,106}
\definecolor{dai_petrol40K}{RGB}{0,67,85}
\definecolor{dai_petrol80}{RGB}{0,122,147}
\definecolor{dai_petrol60}{RGB}{80,151,171}
\definecolor{dai_petrol40}{RGB}{121,174,191}
\definecolor{dai_petrol20}{RGB}{166,202,216}

\definecolor{dai_deepred}{RGB}{113,24,12}
\definecolor{dai_deepred20K}{RGB}{90,19,10}
\definecolor{dai_deepred40K}{RGB}{68,14,7}

\definecolor{apfelgruen}{RGB}{140, 198, 62}
\definecolor{orange}{RGB}{244, 111, 33}
\definecolor{anthrazit}{RGB}{19, 31, 31}

\IEEEoverridecommandlockouts
\begin{document}

\title{\LARGE \bf  A Benchmark for Lidar Sensors in Fog: Is Detection Breaking Down?}

\author{Mario Bijelic$^{*}$, Tobias Gruber$^{*}$ and Werner Ritter%
	\thanks{All authors are with Daimler AG, Wilhelm-Runge-Str. 11, 89081 Ulm, Germany. \{mario.bijelic, tobias.gruber, werner.r.ritter\}@daimler.com}
	\thanks{$^{*}$ Mario Bijelic and Tobias Gruber have contributed equally to the work.}
}

\maketitle

\thispagestyle{empty}
\pagestyle{empty}

\begin{acronym}
 \acro{CMOS}{complementary metal-oxide semiconductor}
 \acro{EU}{European Union}
 \acro{DENSE}{aDverse wEather eNvironment Sensing systEm}
 \acro{FIR}{far infrared} 
 \acro{NIR}{near infrared}
 \acro{SWIR}{short wave infrared}
 \acro{ADAS}{automotive drive assistance system}
 \acro{RMS}{root mean squared}
 \acro{LIDAR}[lidar]{light detection and ranging}
 \acro{TOF}{time of flight}
 \acro{NFIR}{near field infrared}
 \acro{FOV}{field of view}
 \acro{OPAL}{obscurant penetrating autosynchronous lidar}
 \acro{InGaAs}{indium gallium arsenide}
\end{acronym}

\begin{abstract}
Autonomous driving at level five does not only means self-driving in the sunshine. 
Adverse weather is especially critical because fog, rain, and snow degrade the perception of the environment. 
In this work, current state of the art light detection and ranging (lidar) sensors are tested in controlled conditions in a fog chamber. 
We present current problems and disturbance patterns for four different state of the art lidar systems.
Moreover, we investigate how tuning internal parameters can improve their performance in bad weather situations.
This is of great importance because most state of the art detection algorithms are based on undisturbed lidar data.
\end{abstract}

\section{Introduction}

\begin{figure}[t]
	\centering 
	\includegraphics[width=0.95\columnwidth]{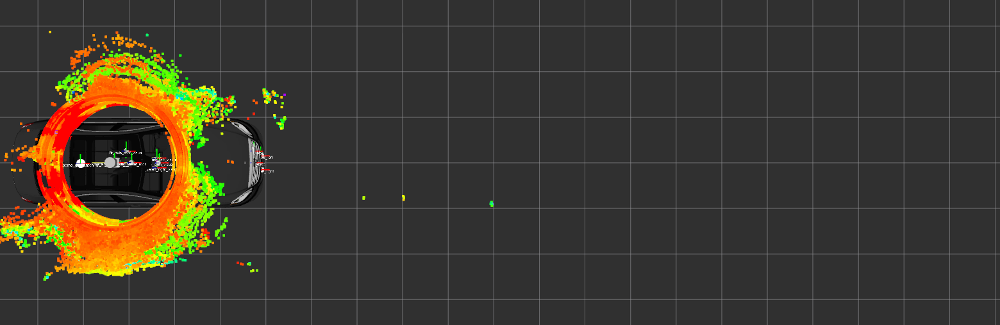}
	\caption{Point cloud of a Velodyne HDL64-S3D in dense fog.}
	\label{fig:title}	
\end{figure}

The race to create smarter sensors for autonomous driving has started. For greater functional safety, various companies are working to develop a third sensor besides camera and radar, to complement their abilities in difficult situations.
Currently, \ac{LIDAR} sensors, gated and \ac{TOF} cameras are competing against each other.
Looking at current proposed sensor setups from established car manufactures and large American IT companies, one can say that \ac{LIDAR} technology has an advantage over its competitors \ac{TOF} and gated imaging.
However, autonomous vehicles at level five require environment perception at all times, even under adverse weather conditions such as fog, haze, mist and rain. 
As Fig.~\ref{fig:title} shows, current \ac{LIDAR} technologies have major problems in scattering media such as fog because they usually work at the limit of eye safety regulations.
Tests in well defined conditions are essential in order to closely investigate the behavior of these sensors in such conditions.

\subsection{Motivation}

How does \ac{LIDAR} work?
Basically, \ac{LIDAR} sends out a laser pulse which is registered at departure and the time is measured until an echo from a target is recorded. 
In detail, the task gets more and more complex as the needed time resolution is high and the maximal signal to noise ratio has to be low. 
Technologically, there are different types of \ac{LIDAR} systems. 
There are \ac{LIDAR} systems for the automotive sector, military, robotics, surveillance or airborne laser scanners for topological mapping.
The detailed solutions differ because the application field is wide. In the automotive sector cost sensitive, fast and high resolution laser scanners are required. The best  known producers are probably Velodyne, Ibeo and Valeo.
State of the art systems are based on mechanical beam steering, optical laser diodes for pulse emission and avalanche photo diodes for detection \cite{IBEOroadref,IBEOLaserscannerPatentboehlau2004optoelektronische, VelodynePatent}.
But current trends \cite{MEMSReviewLidar} indicate the mechanical parts will be replaced by solid state technology which enables more robust and maintainable laser scanners.

The most advanced laser and technologically appealing scanners are the publicly presented military grade products. 
One of those products is the \ac{OPAL} Neptec scanner \cite{Trickey2013,OPALObscurantConditions}. 
In contrast to current automotive sensors, which are based on \unit[905]{nm} lasers, it is based on a \unit[1550]{nm} laser source. 
Also airborne and surveillance laser scanners are based on the \ac{SWIR} region, e.g. the products from Teledyne Optec. 
Unfortunately, this shift to a higher wavelength requires more expensive fiber lasers and \ac{InGaAs} detectors.

In order to compare these different forms of lidar systems in adverse weather conditions, an easy and reproducible benchmark is needed.
Within the \ac{EU} project \ac{DENSE}, we present such a test procedure for state of the art lidar sensors and the results.
Moreover, we show and evaluate disturbance patterns. 
All these results mark a baseline for upcoming developments in the project.

\subsection{Related Work}

\subsubsection{Lidar characterization}

In addition to theoretical characterization of \ac{LIDAR} systems as in \cite{Pratt1969}, experimental evaluation and especially benchmarks between different sensor technologies are very important. 
A first experimental study was already performed in 1991 by Hebert and Krothov \cite{Hebert1992}. 
They verified measurement models with experiments and quantized both range precision and angular precision. 
As soon as 2D laser scanners such as the Sick LMS 200 were commercially available, they were characterized by Ye and Borenstein \cite{Ye2002} through extensive experimentation. 
A square target was precisely moved in front of the laser scanner in order to obtain insight into phenomena such as the range drift over time, the effect of the incidence angle and the mixed pixels problem. 
In 2008, a first assessment of four different laser range finders (Sick LMS200, Hokuyo URG-04LX, Ifm Effector O1D100 and Sick DT60) were performed \cite{Pascoal2008}. 
This benchmark was extended in \cite{Wong2011} where a total of 10 different range sensors were evaluated. In addition to characterization in the lab, measurements in underground environments showed the performance of these sensors in real environments. 

\subsubsection{Influence of bad weather on \ac{LIDAR} systems}

When Peynot et al. recorded a dataset for an advanced perception system for autonomous vehicles in environmental conditions (dust, smoke and rain), they observed that dust particles in the air are detected by laser sensors and hide obstacles behind the dust cloud \cite{Peynot2009}. 
In \cite{Rasshofer2011}, Rasshofer et al. theoretically investigated the influence of weather phenomena based on Mie's theory in order to develop a novel electro-optical laser radar target simulator system. 
They verified their models with real world measurements in fog, rain and snow, however, these measurements rely on random weather conditions and cover only rough visibility differences.
Ijaz et al. showed in \cite{Ijaz2012} that the attenuation due to fog and smoke in a free space optical communication link are not wavelength dependent for visibilities less than \unit[500]{m} and wavelengths in between \unit[600-1500]{nm}.
Their evaluations were performed in a very small laboratory controlled atmospheric chamber.
Trickey et al. investigated the penetration performance of the Opal Lidar in dust, fog, snow whiteouts and smoke \cite{Trickey2013}. 
Dust and fog performances have been evaluated in an aerosol chamber while snow conditions and smoke were obtained in field tests.
A very detailed inspection and discussion of two popular wavelengths for \ac{LIDAR} systems, i.e. \unit[905]{nm} and \unit[1550]{nm}, was done by Wojtanowski et al. \cite{Wojtanowski2014}.
Recently, Hasirlioglu et al. studied the disturbances due to low ambient temperatures and exhaust gases \cite{EffectsExhaustHasirlioglu2017}. 
Considering all these references \cite{Peynot2009,Rasshofer2011,Ijaz2012,Trickey2013,Wojtanowski2014,EffectsExhaustHasirlioglu2017}, it is clear that there is a huge impact from adverse weather on \ac{LIDAR} systems. 
Some even have floated the idea to estimate the weather conditions by comparing \ac{LIDAR} points in good weather and in adverse weather \cite{Zhu2015}.
For autonomous driving in adverse weather, it is very important to gain even more insight into the degradation of \ac{LIDAR} systems. 
This work presents a very detailed performance evaluation of four popular automotive \ac{LIDAR} systems under defined weather conditions in a fog chamber.
The size of the fog chamber allows us to set up a realistic street scenario and it is possible to generate fog with two different droplet distributions and change the visibility in very small steps.

\section{Tested laser scanners}

\begin{table}[t]
	\centering
	\caption{Technical summary of the tested laser scanners. Differences between similar models are marked \textcolor{dai_deepred}{red}.}
	\begin{tabular}{l || c  c  }
		\hline
		Manufacturer                    & Velodyne                                 & Ibeo \\
		Model                           & HDL-64S2/\textcolor{dai_deepred}{S3D}    & LUX /\textcolor{dai_deepred}{LUX HD} \\
		\hline
		Laser layers                    & 64                                       & 4 \\
		Range [m]                       & 50/120                                   & 50/200 \textcolor{dai_deepred}{30/120}\\
		Multiple echoes                      & 1 \textcolor{dai_deepred}{2}             & 3  \\
		Heavy duty                      & No                                       & No \textcolor{dai_deepred}{Yes}\\
		Wavelength [nm]                 & 905                                      & 905\\
		Data rate [Hz]                  & 5-20                                     & 12.5/25/50 \\
		Horiz. \acs{FOV} [$^\circ$]     & 360                                 & 85 \\  
		Horiz. resolution [$^\circ$]         & 0.0864-0.1728                            &  0.125/0.25/0.5 \\
		Vert. \acs{FOV}                  &  26.33                                   & 3.2 \\
		Vert. spacing                    & $1/2$ to $1/3^\circ$                     & 0.8 \\
		Embedded software               & No                                       & Yes\\
		Distance accuracy [cm]        & $\leq$2@1$\sigma$                        & 4 \\
		Operating temp. [$^\circ$C]    & -10 to 50 \textcolor{dai_deepred}{ 60}  & -40 to 85 \\
		Points per second              & $\approx 1.3\times 10^6$ \textcolor{dai_deepred}{ $\approx 2\times 10^6$} & $\approx 44\times 10^3$\\
		\hline
		
	\end{tabular} 
	\label{tab:LaserSpecs}
\end{table}

\begin{figure}
	\begin{tabular}{c  c  c}
		S2 & S2/S3D & S3D\\
		long exp. & short exp. & long exp.\\
		\includegraphics[clip,trim=100 0 100 0,width=2.5cm, height=2.5cm]{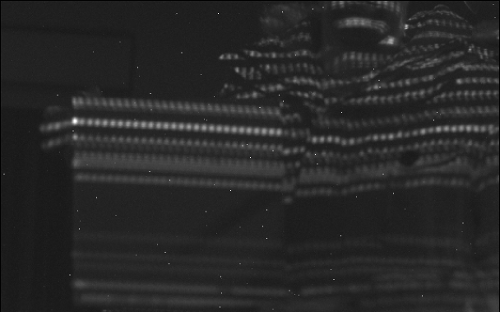} &
		\includegraphics[clip,trim=100 0 100 0,width=2.5cm, height=2.5cm]{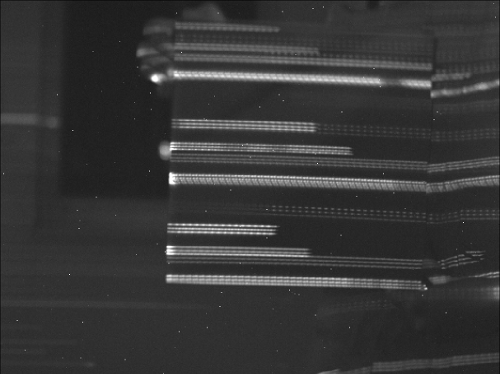} &
		\includegraphics[clip,trim=100 0 100 0,width=2.5cm, height=2.5cm]{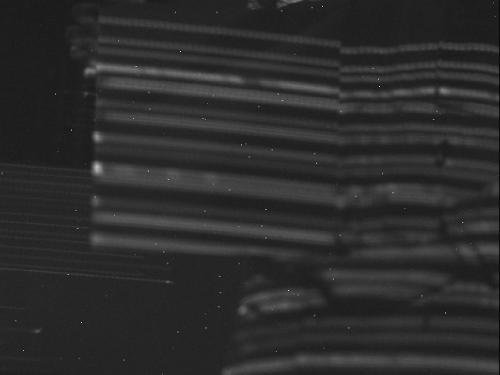}
	\end{tabular}
	\caption{Captured laser intensities with a Raptor OWL-640 camera on a 50\% dry reflective target. A short exposure was set to capture the scan pattern for the S3D and S2, which are equivalent. Longer exposure times illustrate the different scan qualities between the S2 and S3. Notice the increased uniformity for the S3D.}
	\label{fig:LaserIntDistribution}
\end{figure}

In this test, four state of the art laser scanners from Ibeo and Velodyne are compared with one another, i.e. Velodyne HDL-64S2, Velodyne HDL-64S3, Ibeo LUX and Ibeo LUX HD. 
A technical summary of these models can be found in Table~\ref{tab:LaserSpecs}. 
Both Velodyne laser scanners S2 and S3 do not differ in their specifications. 
A 10\,\% reflective target is visible up to \unit[50]{m}. 
A high reflective target with 80\,\% reflectivity is visible up to \unit[120]{m}. 
However, the S2 is an older model and since scan quality has increased (Fig. \ref{fig:LaserIntDistribution}). 
Primarily the intensity calibration was enhanced and is now stable across different scanning layers.
The S3 generation is split into two subproducts: The Velodyne S3S and S3D. 
The S stands for single return and D for dual return. 
In our case, the dual return option is more interesting, as it is able to detect the strongest and last return from a laser pulse and transmit both to a host computer. 
As the bandwidth is limited to 1 Gigabit per second, the horizontal resolution is reduced by approximately 30\,\%.
Counting both strongest and last returned points, the overall number of points increases from $\approx 1.3$ Mio. to $\approx 2$ Mio. points.
We expect that the usage of multiple echoes will enable greater fog penetration.

Furthermore, both Ibeo scanners are technically similar because the Ibeo LUX HD is based on the Ibeo LUX. 
HD designates heavy duty because the LUX HD was originally developed for combine harvesters, where the environment is very dusty. 
The main difference is a modified lens and a tinted covering. 
The sensor was modified in order to enable more robust detections at close distances, as the laser scanners had to face a great deal of whirled-up dust. 
Due to the changes, the maximum viewing distance has been lowered. 
Both laser scanners offer four scanning rows and the same resolution.

In total, the Velodyne and Ibeo scanners differ quite a lot despite both having the same wavelength (\unit[905]{nm}) and mechanical beam steering. The Velodyne scanner offers a 360$^\circ$ \ac{FOV}, while the field of view of the Ibeo scanners is limited to 85$^\circ$. 
Moreover, the resolution of the Velodyne sensors is several magnitudes larger and enables a more detailed environment perception. 
In addition, the Velodyne scanner is able to provide the intensity for each perceived point.

\subsection{Intensity calculation}
For current automotive application, the measured intensity is becoming increasingly important. 
\cite{ReviewRadiometricProcessing} reviews an overall calibration process and possible influences on intensity measures. 
In addition, four levels of intensity calibration are defined. 
It starts with level 0, where raw intensity measures are provided by a laser scanner and ends with level 3, where a rigorous calibration has been applied and the intensity measures are correct and identical across different systems.  
Level 0 calibration methods are provided by the laser scanner manufacturer and the produced output intensities are not comparable to other manufacturers. 

The process of measuring the true intensity differs from each manufacturer and is proprietary, but this measuring process is crucial for a robust environment perception and can increase the robustness in adverse weather, as \cite{kramper2016method} and \cite{WaveformAnalysisUllrich} show.
Only the Velodyne scanners (S2 and S3D) provide a raw intensity measure (level 0), which will be analyzed for possible influences due to fog.

In \cite{hall2011high}, Velodyne claims to increase the capabilities in adverse weather, due to the usage of a digital signal processor which enables dynamically tuning the laser power if a clear object reflection is not obtained from a certain laser pulse.
Both Velodyne S2 and S3D are able to apply this process automatically and to choose from eight power levels. 
The power level is set in an intelligent manner and enables different power levels for different reflective targets within one scan. 
The S3D models has the unique ability to set this parameter manually for all points and return the actual laser power for each laser point, however, at the expense of \unit[3]{bits} meaning a loss of \unit[1.4]{cm} depth resolution. 
Furthermore, the noise level can be set. 
The noise level describes the ground uncertainty floor for a detected signal from spurious light and thermal excitations. 
In this sense, the parameters will be tuned and an optimal performance point will be derived in Section \ref{sec:ParameterEval}.

In order to achieve a level 0 raw data point cloud the hardware inhomogeneity between each laser inside the Velodyne laser scanners has to be corrected. Each laser beam is collimated by an independent lens and driven by an independent laser diode. Therefore, the beam divergence is not uniform. Velodyne provides a factory-side calibration containing a slope $s$ and focal distance $f_d$ for each laser in order to correct the beam divergence. The raw intensity $I_R$ is corrected to $I_C$ with the know distance $d$ from the point measurement by 
\begin{equation}
I_C = I_R + s \left|\left(f_0-\left(1-\frac{d}{\max(d)}\right)^2\right)\right| 
\end{equation}
where $f_0$ is given by
\begin{equation}
f_0 = \left(1-\frac{f_d}{\text{scale}}\right)^2 .
\end{equation}
One has to keep in mind that beam divergence is worse if the laser beam is confronted with a scattering medium and more sophisticated methods can be defined for a known visibility. \cite{DynamicScattering} offers a good introduction into the physics of scattering processes.

\section{Experiments}

Well-defined weather conditions are essential for testing and comparing different sensors. 
To the authors' best knowledge, the climate chamber of CEREMA in Clermont Ferrand (France) is the only one in Europe that is able to produce stable fog with a certain meteorological visual range $V$ in two different fog droplet distributions, i.e. radiation fog (small droplets) and advection fog (large droplets) \cite{Colomb2008}. The fog density is continuously controlled by a transmissometer that measures the atmospheric transmission factor $T$. As described in \cite{Colomb2008}, the meteorological visual range $V$, also known as visibility, can be derived from $T$ as a more practical parameter to determine the fog density. 

\begin{figure}
 \centering
 \includegraphics[width=0.95\columnwidth]{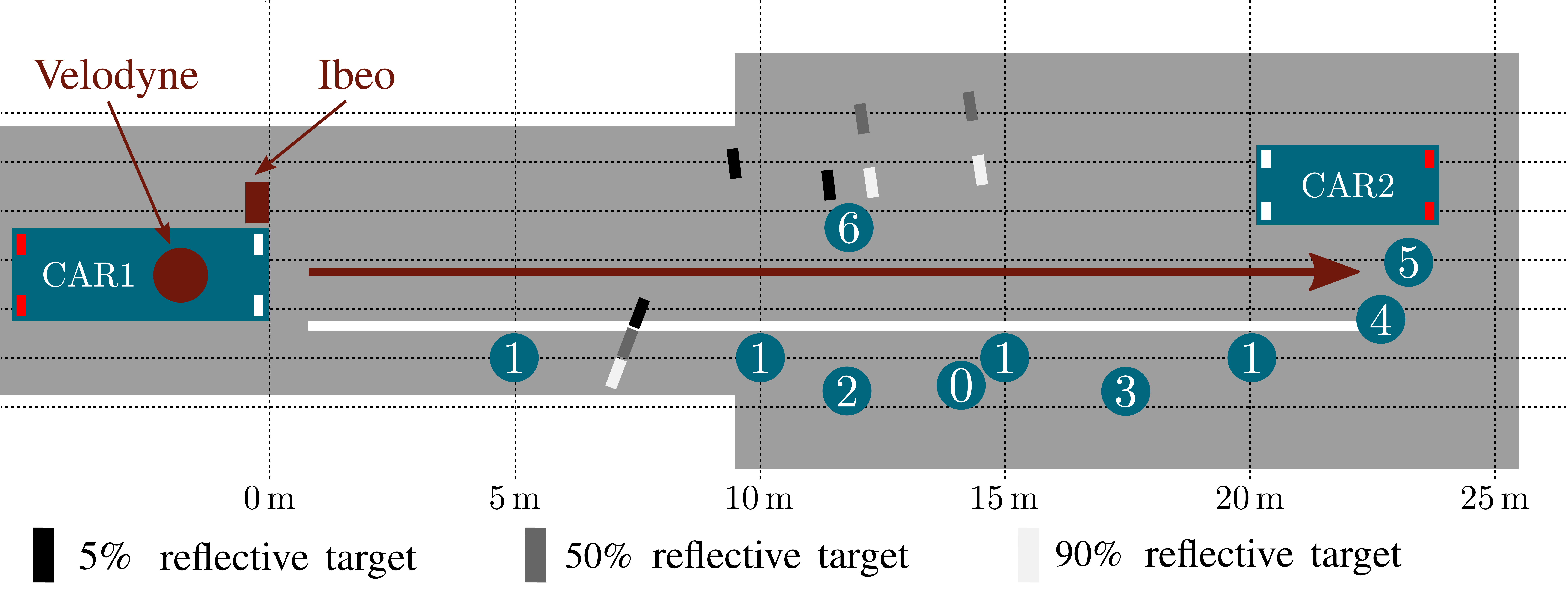}
 \caption{Experiment setup in the fog chamber.}
 \label{fig:experiment_setup}
\end{figure}

\begin{figure}
 \centering
 \includegraphics[width=0.95\columnwidth]{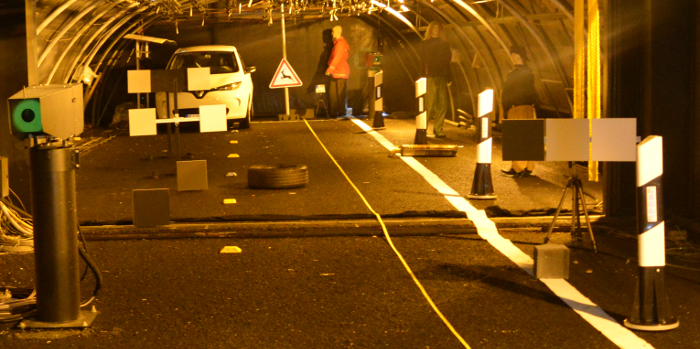}
 \caption{Experiment setup in the fog chamber.}
 \label{fig:experiment_setup_image}
\end{figure}

We set up a static scene as depicted in Fig.~\ref{fig:experiment_setup}. 
The static scene contains many different targets, see Fig.~\ref{fig:experiment_setup_image}: old exhaust pipe (0), reflector posts (1), differently pedestrian mannequins, i.e. child (2), woman (3), man (4), traffic sign (5), tire without rim (6), car (CAR2), road marking and multiple reflective targets.
The \ac{LIDAR} sensors are placed at the beginning of the chamber where the test vehicle (CAR1) is located. 
The Velodyne sensors, i.e. HDL64-S2 and HDL64-S3D, are mounted on top of CAR1 while the Ibeo sensors, i.e. Lux and Lux HD, are mounted as they would be on a bumper on the left side of CAR1 at a height of approximately \unit[0.9]{m} with a forward-looking field of view (see Fig.~\ref{fig:experiment_setup}).
The static scene is recorded under different fog densities and fog droplet distributions in order to investigate the influence of fog to the \ac{LIDAR} performance. 
Because the Velodyne HDL64-S3D internal parameters \emph{laser output power} and \emph{noise ground level} can be varied, the influence of these parameters on measurements in adverse weather can be estimated. 

\begin{figure*}[t]
 \centering 
 
\begin{tabular}{>{\centering\arraybackslash}m{0.5cm} >{\centering\arraybackslash}m{0.2cm} 
>{\centering\arraybackslash}m{0.28\textwidth} >{\centering\arraybackslash}m{ 
0.28\textwidth} >{\centering\arraybackslash}m{0.28\textwidth}}
& &
HDL64-S2 & 
HDL64-S3 strongest & 
HDL64-S3 last \\

\rotatebox[origin=l]{90}{reference} & &
\includegraphics[width=0.25\textwidth]{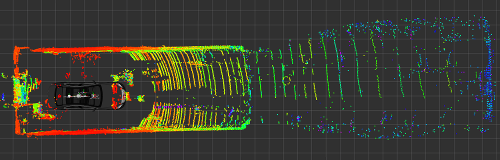} 
& 
\includegraphics[width=0.25\textwidth]{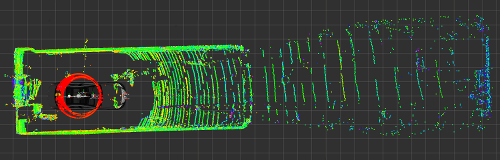} 
&
\includegraphics[width=0.25\textwidth]{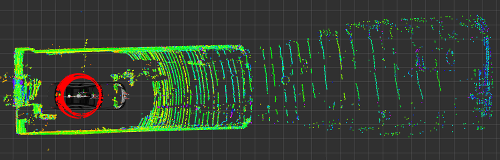} 
\\ 

\multirow{2}{*}{\rotatebox{90}{\parbox[c]{2.7cm}{\centering $V = \unit[45]{m}$}}} &
\rotatebox{90}{\footnotesize radiation} &
\includegraphics[width=0.25\textwidth]{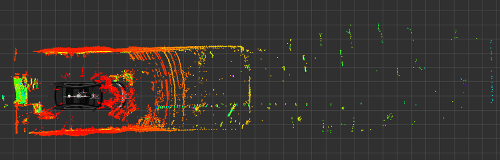} 
& 
\includegraphics[width=0.25\textwidth]{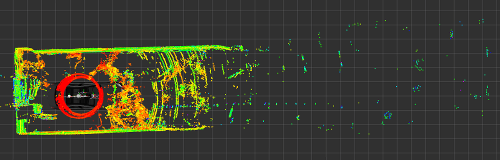} 
&
\includegraphics[width=0.25\textwidth]{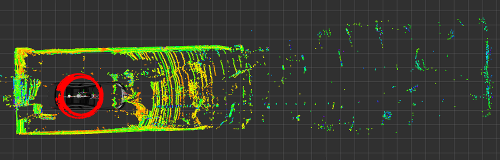} 
\\

&
\rotatebox{90}{\footnotesize advection} & 
\includegraphics[width=0.25\textwidth]{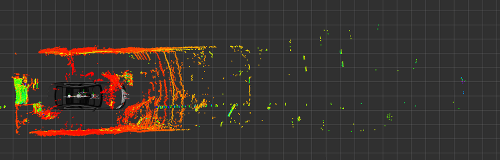} 
& 
\includegraphics[width=0.25\textwidth]{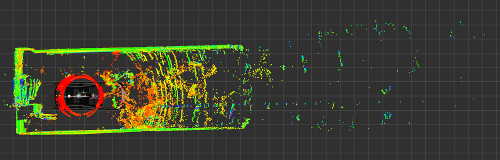} 
&
\includegraphics[width=0.25\textwidth]{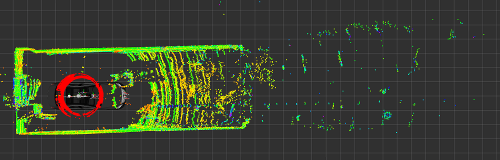} 
\\ 

\multirow{2}{*}{\rotatebox{90}{\parbox[c]{2.7cm}{\centering $V = \unit[36]{m}$}}} &
\rotatebox{90}{\footnotesize radiation} &
\includegraphics[width=0.25\textwidth]{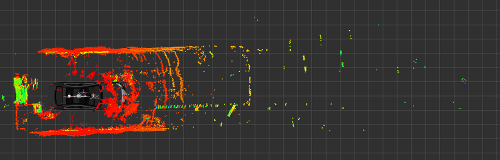} 
& 
\includegraphics[width=0.25\textwidth]{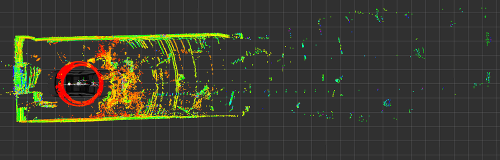} 
&
\includegraphics[width=0.25\textwidth]{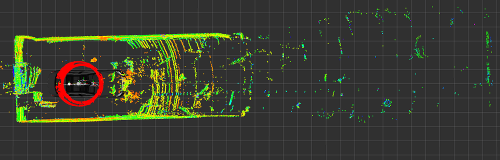} 
\\
 & 
\rotatebox{90}{\footnotesize advection} & 
\includegraphics[width=0.25\textwidth]{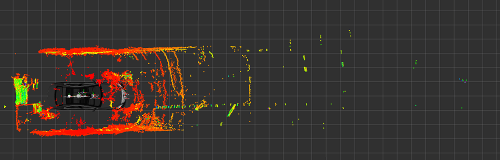} 
& 
\includegraphics[width=0.25\textwidth]{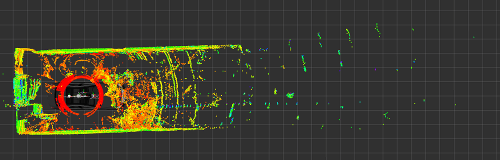} 
&
\includegraphics[width=0.25\textwidth]{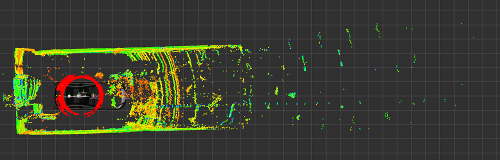} 
\\ 

\multirow{2}{*}{\rotatebox{90}{\parbox[c]{2.7cm}{\centering $V = \unit[24]{m}$}}} &
\rotatebox{90}{\footnotesize radiation} &
\includegraphics[width=0.25\textwidth]{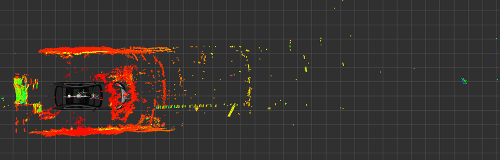} 
& 
\includegraphics[width=0.25\textwidth]{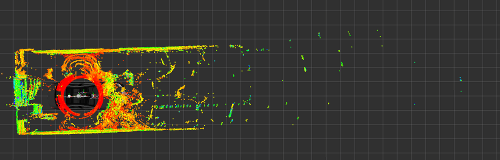} 
&
\includegraphics[width=0.25\textwidth]{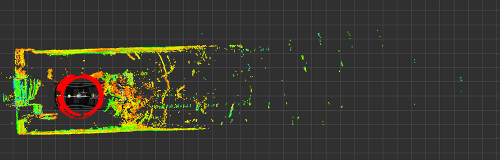} 
\\

& 
\rotatebox{90}{\footnotesize advection} & 
\includegraphics[width=0.25\textwidth]{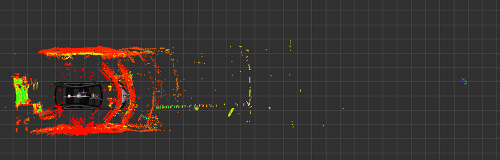} 
& 
\includegraphics[width=0.25\textwidth]{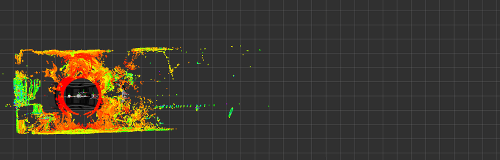} 
&
\includegraphics[width=0.25\textwidth]{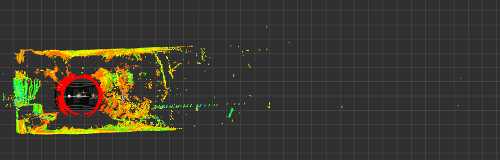} 
\\ 

\multirow{2}{*}{\rotatebox{90}{\parbox[c]{2.7cm}{\centering $V = \unit[13]{m}$}}} &
\rotatebox{90}{\footnotesize radiation} &
\includegraphics[width=0.25\textwidth]{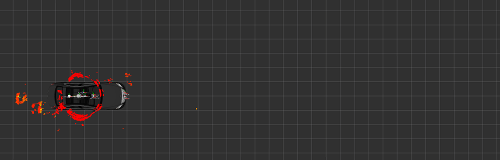} 
& 
\includegraphics[width=0.25\textwidth]{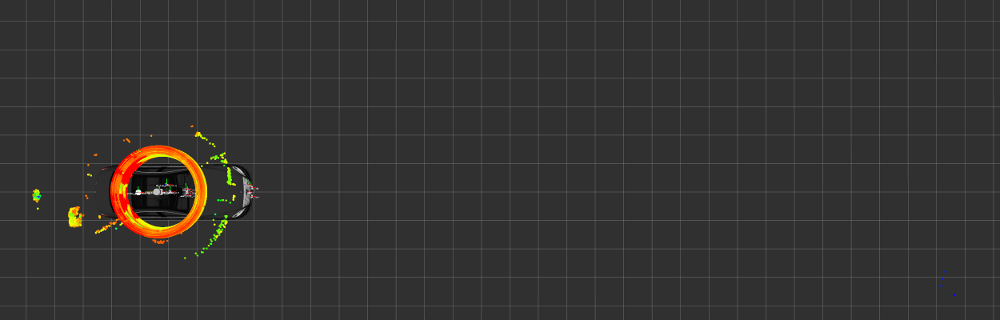} 
&
\includegraphics[width=0.25\textwidth]{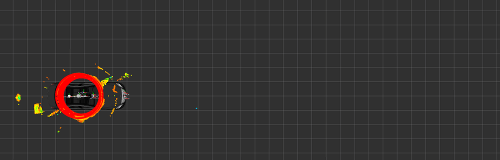} 
\\

& 
\rotatebox{90}{\footnotesize advection} & 
\includegraphics[width=0.25\textwidth]{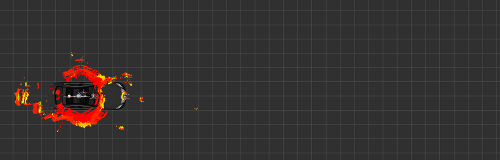} 
& 
\includegraphics[width=0.25\textwidth]{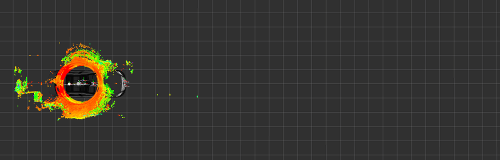} 
&
\includegraphics[width=0.25\textwidth]{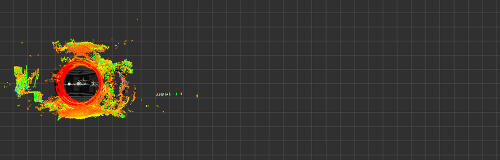} 
\\

 \end{tabular}
 \caption{Bird view of the Velodyne sensors for different fog types and fog densities. Color encodes the intensity.}
 \label{fig:lidar_images_velodyne}
\end{figure*}

\section{Evaluation}
\begin{figure*}[t]
 \centering 
 
\begin{tabular}{>{\centering\arraybackslash}m{0.1cm} 
>{\centering\arraybackslash}m{0.22\textwidth} >{\centering\arraybackslash}m{0.22\textwidth} >{\centering\arraybackslash}m{0.22\textwidth} >{\centering\arraybackslash}m{ 
0.22\textwidth}}

&
\multicolumn{2}{c}{Ibeo LUX} 
&  
\multicolumn{2}{c}{Ibeo LUX HD}  
\\

&
advection 
& 
radiation 
&  
advection 
& 
radiation  
\\

\rotatebox{90}{\parbox[c]{1cm}{\centering reference}}  
& 
\includegraphics[width=0.23\textwidth]{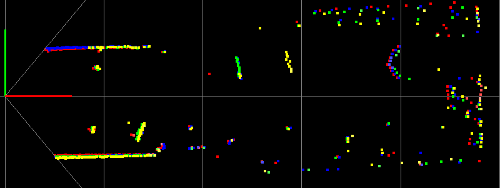} 
& 
\includegraphics[width=0.23\textwidth]{ibeo_topview/clear_lux.PNG} 
&
\includegraphics[width=0.23\textwidth]{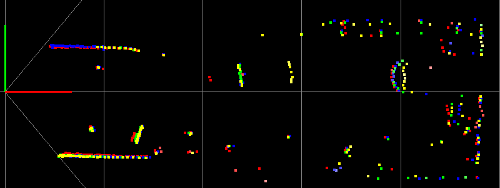} 
& 
\includegraphics[width=0.23\textwidth]{ibeo_topview/clear_luxhd.PNG} 
\\

\rotatebox{90}{\parbox[c]{1cm}{\centering \unit[45]{m}}} 
&
\includegraphics[width=0.23\textwidth]{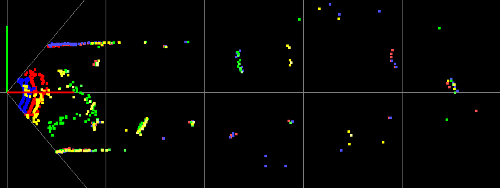} 
& 
\includegraphics[width=0.23\textwidth]{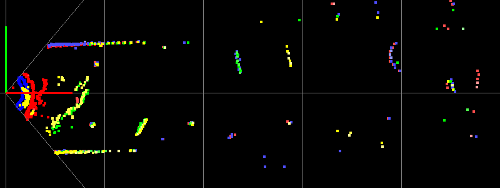} 
&
\includegraphics[width=0.23\textwidth]{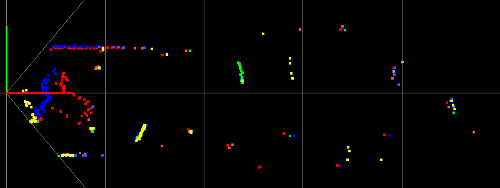} 
& 
\includegraphics[width=0.23\textwidth]{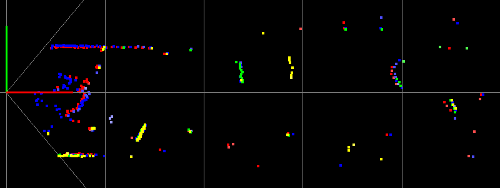} 
\\ 

\rotatebox{90}{\parbox[c]{1cm}{\centering \unit[36]{m}}}
&
\includegraphics[width=0.23\textwidth]{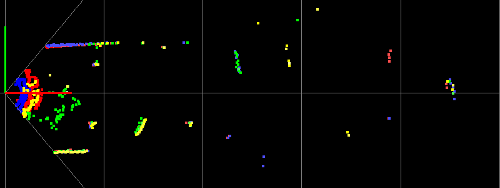} 
& 
\includegraphics[width=0.23\textwidth]{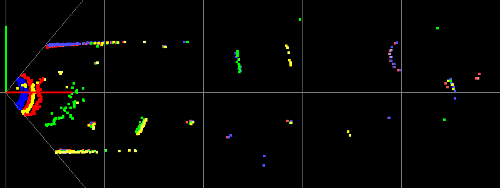} 
& 
\includegraphics[width=0.23\textwidth]{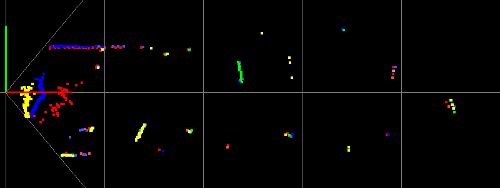} 
& 
\includegraphics[width=0.23\textwidth]{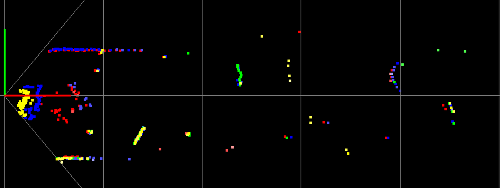} 
\\ 

\rotatebox{90}{\parbox[c]{1cm}{\centering \unit[24]{m}}} 
&
\includegraphics[width=0.23\textwidth]{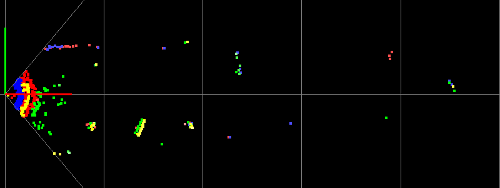} 
& 
\includegraphics[width=0.23\textwidth]{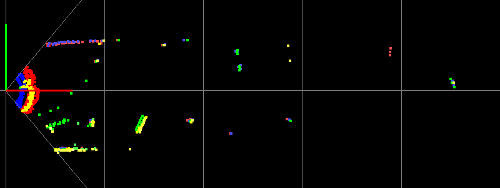} 
& 
\includegraphics[width=0.23\textwidth]{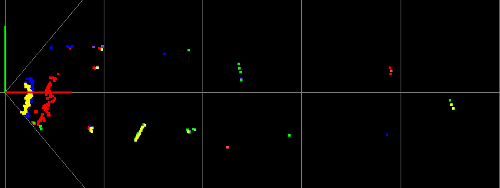} 
& 
\includegraphics[width=0.23\textwidth]{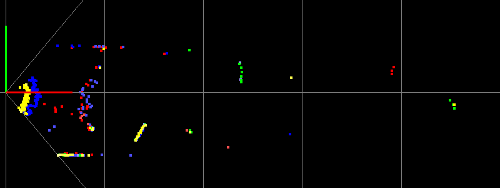} 
\\ 

\rotatebox{90}{\parbox[c]{1cm}{\centering \unit[13]{m}}} 
&
\includegraphics[width=0.23\textwidth]{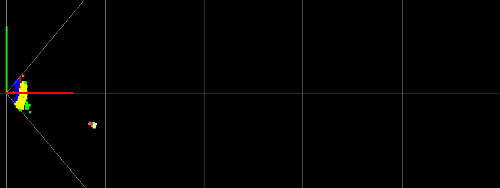} 
& 
\includegraphics[width=0.23\textwidth]{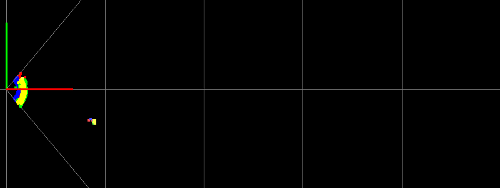} 
& 
\includegraphics[width=0.23\textwidth]{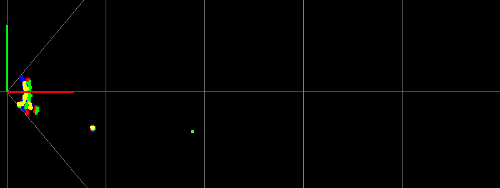} 
& 
\includegraphics[width=0.23\textwidth]{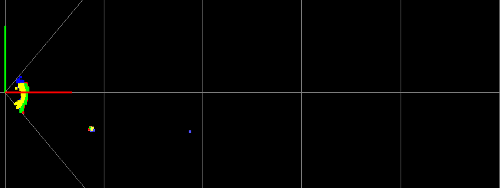} 
\\

 \end{tabular}
 \caption{Bird's-eye views of the Ibeo sensors for different fog types and fog densities. Color encodes the layer: layer 1 (red), layer 2 (blue), layer 3 (green), layer 4 (yellow).}
 \label{fig:lidar_images_ibeo}
\end{figure*}

\subsection{Qualitative Evaluation}

For a first impression of the \ac{LIDAR} performance in fog, we qualitatively assess the bird's-eye views in Fig.~\ref{fig:lidar_images_velodyne} and \ref{fig:lidar_images_ibeo}.
Fig.~\ref{fig:lidar_images_velodyne} shows the bird's-eye views for the strongest and last return for both the Velodyne HDL64-S2 and the HDL64-S3D. The colors encode the intensity.
In general, it is obvious that fog impairs \ac{LIDAR} performance heavily. 
The fog disturbance results in a cloud around the sensor, because the signal is reflected by fog droplets close to the car. 
This creates the need to filter such clutter points, drastically reducing the effective resolution. 
In the HDL64-S3D scans a constant red circle around the laser scanner is visible. Those points appear if the recorded echoes are below the noise level and are therefore projected to their default position. This can only be observed if the power return mode is activated, otherwise no point is returned.
Moreover, the maximal viewing distance is limited by attenuation because the illumination power is limited and possible reflected signals vanish in the signal noise. 
The Velodyne HDL64-S3D delivers better results than the older model HDL64-S2. 
With respect to the bird's-eye view, there is no visible difference between strongest and last return.
For small visibilities around \unit[13]{m}, all sensors suffer more in radiation fog than in advection fog. 
For higher visibilities the performance is equal or even better in radiation fog than in advection fog.

The bird's-eye views for the Ibeo LUX and Ibeo LUX HD are illustrated in Fig.~\ref{fig:lidar_images_ibeo}.
The color coding indicates the corresponding layer, i.e. layer 1 = red, layer 2 = blue , layer 3 = green, layer 4 = yellow. As for the Velodyne sensors, fog reduces the maximal viewing distance heavily. 
By design, the Ibeo LUX HD contains less clutter than the Ibeo LUX. 
Starting from $V =\unit[36]{m}$, both Ibeo sensors are able to detect all targets in the experiment setup while the Velodyne sensors still have difficulties in larger distances. 
There is no difference visible between advection and radiation fog.
Compared to the Velodyne sensors, the Ibeo sensors can perceive the license plate of the car and the traffic sign at a lower visibility starting from \unit[24]{m}.
As there is only a small difference between advection and radiation fog, we continue the following evaluation with only advection fog.

\subsection{Maximum viewing distances}

For a more detailed evaluation, a person successively moved three large well-calibrated Zenith Polymer diffuse reflective targets (\unit[50x50]{cm}) with reflectivities of 5\%, 50\% and 90\% at a height of \unit[0.9]{m} along the main viewing axis (see arrow in Fig.~\ref{fig:experiment_setup}) in order to obtain the maximum viewing distance. 
The person and the target are well separated as the person holds the target \unit[1.5]{m} away from their body with a pole.
The maximum viewing distance is obtained by taking the distance from the last visible point on the targets. 
All sensors used their automatic algorithm to set the proper laser parameters.
Due to the length of the chamber, the evaluation of the maximum viewing distance $d_\text{max}$ is limited.
This limit differs in between the Ibeo and Velodyne sensors because of different mount points.
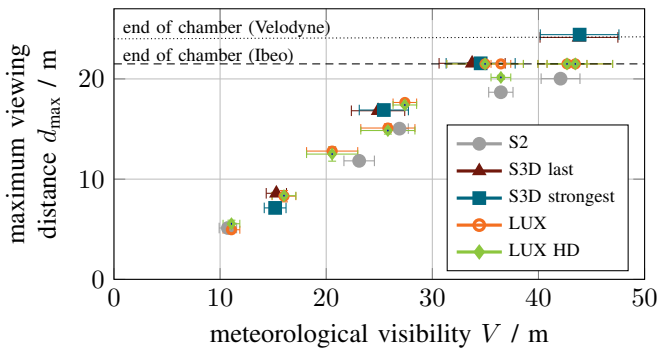
\begin{figure}[t]
  \begin{tikzpicture}
   \begin{axis}[
    xlabel=meteorological visibility $V$ / \unit{m},
    ylabel style={align=center},
    ylabel=maximum viewing\\distance $d_\text{max}$ / \unit{m}, 
    grid=major,
    xmin=0,
    xmax=50,
    ymin=0,
    ymax=27,    
    legend style={
      cells={anchor=west},
      legend pos=south east,
      font=\scriptsize
    },
    legend entries={S2, S3D last, S3D strongest, LUX, LUX HD},
    width=\columnwidth,
    height=0.6\columnwidth
]

\addlegendimage{very thick, dai_ligth_grey40K, mark=*}
\addlegendimage{very thick, dai_deepred, mark=triangle*}
\addlegendimage{very thick, dai_petrol, mark=square*}
\addlegendimage{very thick, orange, mark=o}
\addlegendimage{very thick, apfelgruen, mark=diamond*}

\addplot+ [very thick,
	   solid,
		only marks,
	   color=dai_ligth_grey40K,
	   mark=*,
	   mark size={2}, 
	   mark options={solid,dai_ligth_grey40K}, 
	   error bars/.cd,
	   	   y dir = both, y explicit, 
	   x dir = both, x explicit,  
           error mark options={rotate=90,dai_ligth_grey40K}
]
table[col sep=comma, x index=2, y index=0, x error index=3, y error index=1]{VelodyneData/S2/scenario3max_distance_adv.csv};

\addplot+ [very thick,
	   solid,
		only marks,
	   color=dai_deepred,
	   mark=triangle*,
           mark size={2},
           mark options={solid, dai_deepred}, 
           error bars/.cd,
	   	   y dir = both, y explicit, 
	   	x dir = both, x explicit,  
           error mark options={rotate=90,dai_deepred}
]
table[col sep=comma, x index=2, y index=0, x error index=3, y error index=1]{VelodyneData/S3/last/scenario3max_distance_adv.csv};
   
\addplot+ [very thick,
	   solid,
		only marks,
	   color=dai_petrol,
	   mark=square*,
	   mark size={2}, 
	   mark options={solid,dai_petrol}, 
	   error bars/.cd,
	   	   y dir = both, y explicit, 
	   x dir = both, x explicit,  
           error mark options={rotate=90,dai_petrol}
]
table[col sep=comma, x index=2, y index=0, x error index=3, y error index=1]{VelodyneData/S3/strongest/scenario3max_distance_adv.csv};

    \addplot+ [
			color=orange,
			very thick,
			only marks,
			solid,
		  mark=o,
			mark size={1.5}, 
			error bars/.cd,
      y dir=both,
      y explicit,
			x dir=both,
      x explicit, 
      error mark options={
				rotate=90,
				color=orange,
			}
		]
	  table[col sep=comma, x={mean_vis}, y={lux_mean}, x error = {std_vis}, y error = {lux_std}]{ibeo_lidar_plot/advection_90.txt}; 
			
		\addplot+ [
			color=apfelgruen,
			very thick,
			only marks,
			solid,
		  mark=diamond*,
			mark size={1.5}, 
			error bars/.cd,
      y dir=both,
      y explicit,
			x dir=both,
      x explicit, 
      error mark options={
				rotate=90,
				color=apfelgruen,
			}
		]
	  table[col sep=comma, x={mean_vis}, y={luxhd_mean}, x error = {std_vis}, y error = {luxhd_std}]{ibeo_lidar_plot/advection_90.txt};

\addplot[mark=none, densely dashed, black] coordinates {(0,21.5) (50,21.5)};
\addplot[mark=none, densely dotted, black] coordinates {(0,24) (50,24.2)};

\node[anchor=west] at (axis cs: 0,22.3) {\scriptsize end of chamber (Ibeo)};
\node[anchor=west] at (axis cs: 0,25) {\scriptsize end of chamber (Velodyne)};

  \end{axis}
 \end{tikzpicture}
 \caption{Maximum viewing distance comparison for all sensors in advection fog and reflective targets with 90\% reflectivity. }
\label{fig:maxdistances}
\end{figure}
Fig.~\ref{fig:maxdistances} shows the maximum viewing distance vs meteorological visibility $V$ for all four sensors, based on the 90\,\% target.
Since the targets are kept within the fog chamber and get wet as is common for objects in foggy conditions, they are not ideally diffusive reflective and a specular component can influence the measurements.
In order to keep the targets dry, heating elements would be required and these would influence the fog homogeneity.
Keeping the fog homogeneous is more important and as the targets are always directed towards the laser scanner, our results show an upper bound which can decrease for dry targets. 
Fig.~\ref{fig:maxdistances} clearly illustrates that fog limits the maximum viewing distance $d_\text{max}$. 
At fog densities with $V < \unit[40]{m}$, $d_\text{max}$ is significantly less than \unit[25]{m}.
This is critical for driver safety as weather conditions can drop below \unit[50]{m} visibility and the perceived depth from the laser scanners does not allow emergency stops on time. All sensors show equal performance within \unit[0.5]{m}, despite the older Velodyne HDL64-S2.
In this evaluation, there is almost no difference between strongest and last return. This is probably due to the specular component. Therefore, the last returned point is also the strongest.

\subsection{Strongest Return vs. Last Return}

Strongest and last return differ in modality. In general, the strongest return offers a more stable and accurate measure. But in adverse weather more received pulses grant a higher probability to get an actual object responses. 
This comparison points out the difference between different echoes in fog. 
In order to elaborate this difference, the static scenario is evaluated as follows: the point cloud \unit[1.5]{m} in front of the car is binned in 122 depth slices of \unit[0.25]{m}. 
Then, the number of points in each depth bin is obtained by averaging 100 point clouds.
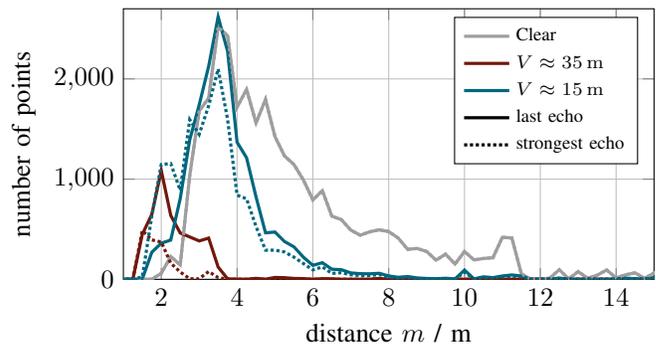
\begin{figure}[t]
	\begin{tikzpicture}
	\begin{axis}[
	xlabel=distance $m$ / \unit{m},
	ylabel=number of points,
	grid=major,
	xmin=1,
	xmax=15,
	ymin=0,
	ymax=2700,    
	legend style={
		cells={anchor=west},
		legend pos=north east,
		font=\scriptsize
	},
	legend entries={Clear, $V \approx \unit[35]{m}$, $V \approx \unit[15]{m}$, last echo, strongest echo},
	width=\columnwidth,
	height=0.6\columnwidth
	]
	
	\addlegendimage{very thick, dai_ligth_grey40K}
	\addlegendimage{very thick, dai_deepred}
	\addlegendimage{very thick, dai_petrol}
	\addlegendimage{very thick, black, solid}
	\addlegendimage{very thick, black, densely dotted}
	
	\addplot+ [very thick,
	solid,
	color=dai_ligth_grey40K,
	mark=none,
	]
	table[col sep=space, x index=0, y index=9]{VelodyneData/power_eval/averages_last.csv};
	
	\addplot+ [very thick,
	solid,
	color=dai_deepred,
	mark=none,
	]
	table[col sep=space, x index=0, y index=1]{VelodyneData/power_eval/averages_last.csv};
	
	\addplot+ [very thick,
	solid,
	color=dai_petrol,
	mark=none,
	]
	table[col sep=space, x index=0, y index=4]{VelodyneData/power_eval/averages_last.csv};
	
	\addplot+ [very thick,
	densely dotted,
	color=dai_ligth_grey40K,
	mark=none,
	]
	table[col sep=space, x index=0, y index=9]{VelodyneData/power_eval/averages_strong.csv};
	
	\addplot+ [very thick,
	densely dotted,
	color=dai_deepred,
	mark=none,
	]
	table[col sep=space, x index=0, y index=1]{VelodyneData/power_eval/averages_strong.csv};
	
	\addplot+ [very thick,
	densely dotted,
	color=dai_petrol,
	mark=none,
	]
	table[col sep=space, x index=0, y index=4]{VelodyneData/power_eval/averages_strong.csv};
	
	\end{axis}
	\end{tikzpicture}
	\caption{Number of points in distance bins for clear weather and advection fog different visibilities $V$. For each fog density, the strongest and last echo is shown.}
	\label{fig:BinnedDepthStrongvsLast}
\end{figure}
Fig.~\ref{fig:BinnedDepthStrongvsLast} shows the mean number of points obtained in each depth bin for different echoes. The power level is chosen automatically by the laser scanner and the noise level is set to the standard value 24.
For clear weather, the strongest return and last return give the same point cloud which serves as ground truth reference.
By increasing the fog density, it can be clearly seen that the number of points decreases and that the last return shows a larger overlap with the reference point cloud than the strongest return. 
The higher number of points compared to the ground truth is due to the clutter which is received at very close distance.

\input{plot_power_level.tex}
\subsection{Parameter Evaluation}\label{sec:ParameterEval}

As described before, the Velodyne HDL64-S3D offers the possibility to manually adjust the internal parameters laser output power and noise ground level.
The power level can be set to an integer value in between 0 and 7. The noise level is a byte value, thus 0-255, which corresponds to the intensity value range. 
The measurement was iterated over the power levels $1-7$. For each power level the noise levels $\left[0, 12, 24, 48 , 96, 192, 240 \right]$ were set.
Fig.~\ref{fig:lidar_power_level} shows the number of points and intensities binned in depth slices over the measurement time in advection fog and for the last echo.
As already mentioned, there is a clear correlation between the fog density and the number of points or rather the intensities.
For all measurements, it can be seen that higher laser power guarantee a longer range while the noise level can adjust possible clutter. 
\begin{figure}[t]
  \begin{tikzpicture}
   \begin{axis}[
    ylabel=number of points,
    xlabel=laser power level,
    grid=major,
    xmin=1,
    xmax=7,
    ymin=0,
    ymax=35000, 
	scaled y ticks = false, 
    legend style={
      cells={anchor=west},
      legend pos=north west,
      font=\scriptsize
    },
    legend entries={Clear, $V\approx\unit[15]{m}$, $V\approx\unit[25]{m}$ , $V\approx\unit[35]{m}$, $V\approx\unit[45]{m}$},
    width=\columnwidth,
    height=0.6\columnwidth
]

\addlegendimage{very thick, dai_ligth_grey40K, mark=none}
\addlegendimage{very thick, dai_deepred, mark=none}
\addlegendimage{very thick, dai_petrol, mark=none}
\addlegendimage{very thick, orange, mark=none}
\addlegendimage{very thick, apfelgruen, mark=none}

\addplot+ [very thick,
	   solid,
	   color=dai_ligth_grey40K,
	   mark=none,
]
table[col sep=comma, x index=0, y index=1]{VelodyneData/power_eval/last/Clear_1500number_max.csv};

\addplot+ [very thick,
	   solid,
	   color=dai_deepred,
	   mark=none,
]
table[col sep=comma, x index=0, y index=1]{VelodyneData/power_eval/last/Adv_10number_max.csv};


\addplot+ [very thick,
	   solid,
	   color=dai_petrol,
	   mark=none,
]
table[col sep=comma, x index=0, y index=1]{VelodyneData/power_eval/last/Adv_20number_max.csv};


\addplot+ [very thick,
	   solid,
	   color=orange,
	   mark=none,
]
table[col sep=comma, x index=0, y index=1]{VelodyneData/power_eval/last/Adv_30number_max.csv};


\addplot+ [very thick,
	   solid,
	   color=apfelgruen,
	   mark=none,
]
table[col sep=comma, x index=0, y index=1]{VelodyneData/power_eval/last/Adv_40number_max.csv};


\end{axis}
 \end{tikzpicture}
  \caption{Number of points achieved with a noise level of 0 plotted against the power level.}
\label{fig:ŚhiftPowerLevelPoints}
\end{figure}
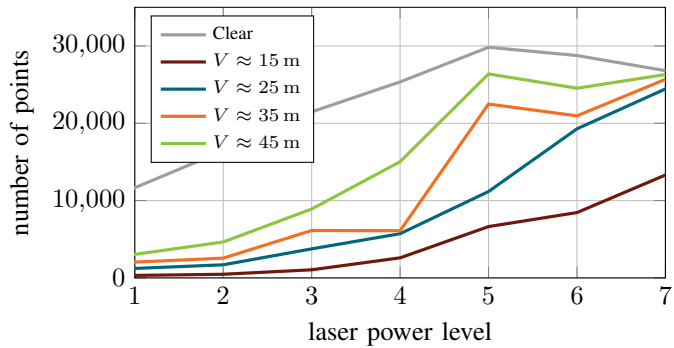
Fig.~\ref{fig:ŚhiftPowerLevelPoints} gives the number of points vs power level at one single noise level 0 for different fog densities. 
For increased fog densities, the number of returned points increases with the laser output power.
However, for clear conditions or light fog, high power levels make the laser prune to oversaturate the detector. 
Increasing the noise level means that more points in close distance can be observed and therefore the clutter points show a high intensity. 
This indicates that a depth-dependent noise level is required to filter out possible clutter points \cite{kramper2016method}. 
However, both parameters are restricted to either saturate the receiver if the laser power is too high, or to neglect many points if the noise level is too high.

\begin{figure}[t]
  \begin{tikzpicture}
   \begin{axis}[
    xlabel=distance $d$ / m,
    ylabel=raw intensity,
    grid=major,
    xmin=3,
    xmax=25,
    ymin=50,
    ymax=150,    
    legend style={
      cells={anchor=west},
      legend pos=north west,
      font=\footnotesize
    },
    legend entries={$V\approx\unit[15]{m}$, 5\,\% target, $V\approx\unit[25]{m}$, 90\,\% target, $V\approx\unit[45]{m}$},
    legend columns=2,
    legend pos= south east,
    width=\columnwidth,
    height=0.65\columnwidth
]

\addlegendimage{dai_petrol, very thick}
\addlegendimage{black, solid, very thick}
\addlegendimage{dai_deepred, very thick}
\addlegendimage{black, densely dotted, very thick}
\addlegendimage{dai_ligth_grey40K, very thick}



\addplot+ [very thick,
	   solid,
	   color=dai_deepred,
	   mark=none,
       error bars/.cd,
	   y dir = both, y explicit, 
	   x dir = both, x explicit,  
       error mark options={rotate=90,dai_deepred}
]
table[col sep=comma, x index=0, y index=1, x error index=3, y error index=4]{VelodyneData/S3/last/Adv/scenario3/binned_data_visi_24.88_std_2.51.csv};

\addplot+ [very thick,
	   solid,
	   color=dai_petrol,
	   mark=none,
       error bars/.cd,
	   y dir = both, y explicit, 
	   x dir = both, x explicit,  
       error mark options={rotate=90,dai_petrol}
]
table[col sep=comma, x index=0, y index=1, x error index=3, y error index=4]{VelodyneData/S3/last/Adv/scenario3/binned_data_visi_15.30_std_0.95.csv};

\addplot+ [very thick,
	   solid,
	   color=dai_ligth_grey40K,
	   mark=none,
       error bars/.cd,
	   y dir = both, y explicit, 
	   x dir = both, x explicit,  
       error mark options={rotate=90,dai_ligth_grey40K}
]
table[col sep=comma, x index=0, y index=1, x error index=3, y error index=4]{VelodyneData/S3/last/Adv/scenario3/binned_data_visi_43.82_std_3.67.csv};


\addplot+ [very thick,
	   densely dotted,
	   color=dai_deepred,
	   mark=none,
       error bars/.cd,
	   y dir = both, y explicit, 
	   x dir = both, x explicit,  
       error mark options={rotate=90,dai_deepred}
]
table[col sep=comma, x index=0, y index=1, x error index=3, y error index=4]{VelodyneData/S3/last/Adv/scenario1/binned_data_visi_23.33_std_1.19.csv};

\addplot+ [very thick,
  	   densely dotted,
	   color=dai_petrol,
	   mark=none,
       error bars/.cd,
	   y dir = both, y explicit, 
	   x dir = both, x explicit,  
       error mark options={rotate=90,dai_petrol}
]
table[col sep=comma, x index=0, y index=1, x error index=3, y error index=4]{VelodyneData/S3/last/Adv/scenario1/binned_data_visi_15.28_std_0.89.csv};

\addplot+ [very thick,
	   densely dotted,
	   color=dai_ligth_grey40K,
	   mark=none,
       error bars/.cd,
	   y dir = both, y explicit, 
	   x dir = both, x explicit,  
       error mark options={rotate=90,dai_ligth_grey40K}
]
table[col sep=comma, x index=0, y index=1, x error index=3, y error index=4]{VelodyneData/S3/last/Adv/scenario1/binned_data_visi_43.30_std_6.73.csv};

  \end{axis}
 \end{tikzpicture}
 \caption{Raw Intensities against distance of the HDL64-S3D on reflective targets (5\,\% and 90\,\%) for three different fog densities in advection fog.}
 \label{fig:IntensDegSecen1Secen3}
\end{figure}
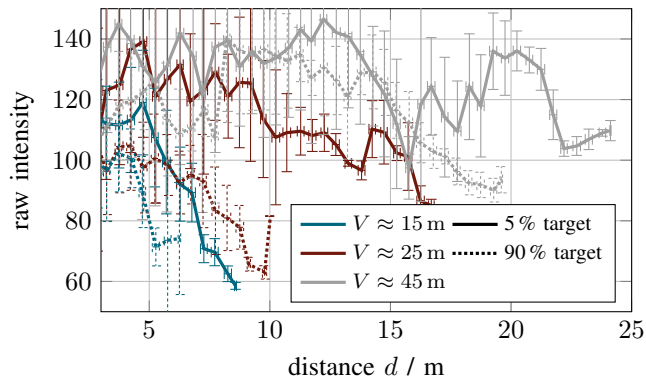
\subsection{Intensity Evaluation}

The intensity values are of interest as they correspond directly to the measured signal peak amplitude. 
Therefore, the intensity also determines when the signal disappears. 
In this sense, it is interesting to determine which intensity is measured. 
In Fig.~\ref{fig:IntensDegSecen1Secen3} the raw intensities for the 5\,\% and 90\,\% target against distance are plotted. 
The HDL64-S3D scanner is set to automatic laser power and the noise level is set to 24. 
As described in \cite{Narasimhan2002}, intensity decreases exponentially due to attenuation. 
However, a fit to an exponential scattering model is not possible under the given confidence as the values jump when a less intense laser ring disappears from the target and is replaced by a higher intense laser ring. 
This inhomogeneity in between the laser rings of the HDL64-S3D can bee seen in Fig.~\ref{fig:LaserIntDistribution} and is worse for the HDL64-S2. 
On average, 5\,\% reflective targets show a $\approx$ 20\,\% lower intensity than 90\,\% targets. 
It is worth to mention that the target disappears before reaching the minimum noise level.

\section{Conclusion and Outlook}
In conclusion, the performance of all four state of the art laser scanners breaks down in fog. 
In direct comparison, only the older Velodyne HDL64-S2 is outperformed. 
Other sensors show similar performance, which is also due to the strict eye safety constraints \ac{LIDAR} systems are facing within the \ac{NIR} band to achieve laser class 1. 

Manually adjusting internal parameters increases performance, e.g. increasing the laser output power for the HDL64-S3D increases range. 
Under adverse weather conditions, the sensors on average run only on level 5 out of 7 possible levels. 
The sensor algorithm may be confused by early high intense clutter responses which lead to a reduction of laser power. 
The authors assume that if any point is measured, the laser power is not increased further. 

Overall in fog the maximal viewing distance is limited to a fraction of the clear weather capabilities. 
Below \unit[40]{m} meteorological visibility, perception is limited to less than \unit[25]{m}. 
Applying multiple echoes and adaptive noise levels increases performance in total by a magnitude of a few meters. 
But this is far away from reliable perception in dense foggy conditions. 
If this problem is not solved, autonomous driving will be limited to level four. 

Therefore, intense research is needed to adapt \ac{LIDAR} solutions from the military or airborne sector to cost sensitive and scalable laser scanners in the \ac{SWIR} region.
Otherwise, the transition to other technologies is interesting, e.g. gated imaging. 
For example, Goodrich and Brightway Vision provide gated cameras which work better under adverse weather but offer lower depth accuracy \cite{Bijelic2018}. 

Further sensor tests have to be applied in order to monitor upcoming adverse weather systems and their improvements. 
This test provides a benchmark for current sensors and creates an awareness for sensor development. 
These findings are also interesting for upcoming raw data sensor fusion object detection systems like in \cite{MV3D}. 
Such methods rely purely on \ac{LIDAR} data as proposal method and are therefore prone to failure with current \ac{LIDAR} sensors in fog.

\section*{Acknowledgment}
The research leading to these results has received funding from the European Union under the H2020 ECSEL Programme as part of the DENSE project, contract number 692449. We gratefully acknowledge the support from Ibeo and CEREMA. The authors thank CEREMA for the test facility and Ibeo for providing the Ibeo LUX/LUX HD. 

\balance
\bibliographystyle{IEEEtran}
\bibliography{lidar_fog_chamber}

\end{document}